\documentclass[11pt]{article}

\usepackage[final]{automl}

\usepackage{natbib}
\usepackage[utf8]{inputenc}
\usepackage[T1]{fontenc}
\usepackage{hyperref}
\usepackage{url}
\usepackage{booktabs}
\usepackage{amsfonts}
\usepackage{nicefrac}
\usepackage{microtype}
\usepackage{xcolor}
\usepackage{graphicx}
\usepackage{subcaption}
\usepackage{longtable}
\usepackage{array}
\usepackage{placeins}
\usepackage{float}

\usepackage{listings}
\usepackage{fvextra}
\usepackage{tcolorbox}
\tcbuselibrary{breakable,skins,listings}

\lstdefinestyle{evoyaml}{
  basicstyle=\ttfamily\scriptsize,
  breaklines=true,
  breakatwhitespace=false,
  columns=fullflexible,
  keepspaces=true,
  showstringspaces=false,
  frame=none,
  numbers=left,
  numberstyle=\tiny\color{gray},
  xleftmargin=1.5em,
  aboveskip=0.5em,
  belowskip=0.5em
}

\newtcblisting{evolisting}{
  breakable,
  enhanced,
  colback=black!1,
  colframe=black!20,
  boxrule=0.4pt,
  arc=2pt,
  outer arc=2pt,
  left=4pt,
  right=4pt,
  top=4pt,
  bottom=4pt,
  listing only,
  listing options={style=evoyaml}
}

\lstdefinestyle{promptstyle}{
  basicstyle=\ttfamily\fontsize{6.5pt}{7.8pt}\selectfont,
  breaklines=true,
  breakatwhitespace=false,
  columns=fullflexible,
  keepspaces=true,
  showstringspaces=false,
  frame=none,
  xleftmargin=0.5em,
  aboveskip=0.3em,
  belowskip=0.3em
}

\newtcblisting{promptbox}[1][]{
  breakable,
  enhanced,
  colback=blue!2,
  colframe=blue!25,
  boxrule=0.4pt,
  arc=2pt,
  outer arc=2pt,
  left=3pt,
  right=3pt,
  top=3pt,
  bottom=3pt,
  listing only,
  listing options={style=promptstyle},
  title={\small\sffamily #1},
  coltitle=blue!50!black,
  fonttitle=\bfseries\small,
  attach boxed title to top left={yshift=-2mm,xshift=4mm},
  boxed title style={colback=blue!10,colframe=blue!25,boxrule=0.3pt}
}

\newtcblisting{toonbox}{
  breakable,
  enhanced,
  colback=green!2,
  colframe=green!30!black,
  boxrule=0.4pt,
  arc=2pt,
  outer arc=2pt,
  left=3pt,
  right=3pt,
  top=3pt,
  bottom=3pt,
  listing only,
  listing options={style=promptstyle}
}

\newtcblisting{memobox}{
  breakable,
  enhanced,
  colback=orange!3,
  colframe=orange!40!black,
  boxrule=0.4pt,
  arc=2pt,
  outer arc=2pt,
  left=3pt,
  right=3pt,
  top=3pt,
  bottom=3pt,
  listing only,
  listing options={style=promptstyle}
}

\title{EvoForest: A Novel Machine-Learning Paradigm via Open-Ended Evolution of Computational Graphs}

\author[1]{\nameemail{Kamer Ali Yuksel}{kamer@aixplain.com}}
\author[1]{\nameemail{Hassan Sawaf}{hassan@aixplain.com}}
\affil[1]{aiXplain, Inc., San Jose, CA, USA}

\hypersetup{%
  pdftitle={EvoForest: A Novel Machine-Learning Paradigm via Open-Ended Evolution of Computational Graphs},
  pdfsubject={AutoML, Evolutionary Computation, Feature Discovery},
  pdfkeywords={evolutionary computation, feature engineering, computational graphs, neuro-symbolic AI}
}

\begin{document}

\maketitle

\begin{abstract}
Modern machine learning is still largely organized around a single recipe: choose a parameterized model family and optimize its weights. This paradigm has been remarkably successful, but it is too narrow for many structured prediction problems, where the main bottleneck is not parameter fitting but discovering what should be computed from the data. Success often depends on identifying the right transformations, statistics, invariances, interaction structures, temporal summaries, gating mechanisms, or nonlinear compositions, especially when the target objective is non-differentiable, evaluation is cross-validation-based, interpretability matters, or continual adaptation is required. In this paper, we present EvoForest, a hybrid neuro-symbolic system for end-to-end open-ended evolution of computation. EvoForest does not merely generate features: it jointly evolves reusable computational structure, callable function families, and trainable low-dimensional continuous components inside a shared directed acyclic graph (DAG), while scoring candidate graph configurations directly against a non-differentiable cross-validation target. Intermediate nodes store alternative implementations, callable nodes encode reusable transformation families such as projections, gates, and activations, output nodes define candidate predictive computations, and persistent global parameters can be refined by gradient descent. For each configuration, EvoForest evaluates the discovered computation graph and uses a lightweight Ridge-based readout to score the resulting representation. Crucially, the downstream evaluator is not only a selector but also a diagnostic instrument: its internals are converted into structured feedback on importance, redundancy, class separation, residual nonlinearity, and stability, which is then summarized into persistent memoranda that guide future LLM-driven mutation proposals. This design yields a fast and resource-efficient way to outperform conventional machine-learning pipelines. EvoForest performs rapid open-ended search over both representation-learning structure and domain-specific computations, while retaining a parameter-efficient final predictor suitable for continual learning and efficient re-optimization under changing data. In the 2025 ADIA Lab Structural Break Challenge, EvoForest reached 94.13\% ROC-AUC after 600 evolution steps, substantially exceeding the publicly reported winning score of 90.14\% under the same evaluation protocol. More broadly, EvoForest serves as a concrete instance of a search-first paradigm for machine learning, in which learning is driven not only by parameter optimization, but by the explicit discovery, reuse, refinement, and selection of useful computations end-to-end.
\end{abstract}

\section{Introduction}

Modern machine learning is still built around a single dominant recipe: choose a parameterized model, define a loss, and optimize its weights. This paradigm has been highly successful, but it is too narrow for many structured prediction problems. In such settings, the main challenge is often not fitting more parameters, but discovering what should be computed from the data. In tabular, time-series, scientific, financial, healthcare, and industrial problems, predictive performance often depends on the right computations: change statistics, interaction terms, temporal summaries, invariances, similarity functions, gating mechanisms, or domain-specific nonlinear transformations. These are often the real substance of the problem. Yet current machine learning systems usually handle them in only two ways: either humans design them manually, which is powerful but expensive and hard to scale, or large differentiable models are expected to discover them implicitly through end-to-end training. Existing approaches only partially address this gap. Automated feature engineering usually relies on fixed transformation libraries. Genetic programming explores richer symbolic spaces, but often with brittle trees and limited computation reuse. Neural architecture search improves model structure, but still focuses on neural networks as the main learned object. Hyperparameter optimization for models such as XGBoost or LightGBM improves fixed learners without changing what is computed. Recent LLM-based feature generation adds flexibility, but usually lacks reusable graph structure, persistent memory, and diagnostic feedback. We argue for a broader perspective: \emph{search-first machine learning}. In this view, the main learned object is not just a set of weights, but an evolving computational structure. The central question is not only \emph{which model should be trained?} but also \emph{which computations should exist?} Learning becomes structured search over reusable computations, while downstream prediction can often be handled by sparse, closed-form, or otherwise parameter-efficient readouts. This does not replace gradient-based learning, but repositions it: gradients refine useful continuous components inside a broader searched structure rather than defining the entire learning process.

We present EvoForest as a concrete instance of this paradigm. EvoForest is a hybrid neuro-symbolic system that performs end-to-end open-ended evolution of computation through a population-encoded directed acyclic graph (DAG). Each node stores multiple alternative implementations. Intermediate nodes represent reusable computations, output nodes define candidate predictive computations, callable nodes define reusable transformation families such as activations, gates, and projections, and a persistent global store contains trainable tensors that can be refined by gradient descent. A graph configuration selects one alternative per reachable intermediate or callable node, evaluates all output computations, and scores the resulting representation with a lightweight Ridge-based readout. The search space therefore spans both domain-specific features and representation-learning structure. EvoForest is therefore not merely an automated feature engineering system. It jointly evolves shallow domain-specific statistics and deeper learnable computational motifs. It can rediscover classical hand-designed transformations, but can also compose them into richer nonlinear pipelines involving callable transforms, learned projections, and shared reusable subgraphs. In this sense, EvoForest performs a form of fast architecture search and autonomous computation discovery at the same time, but over a broader space than standard NAS because the search object is not restricted to neural architectures. A second defining property of EvoForest is that it optimizes directly for non-differentiable evaluation targets. In many practical settings, performance is measured by cross-validation aggregates, ranking metrics, robustness criteria, or other objectives that are not naturally aligned with end-to-end differentiable surrogates. EvoForest is designed for this regime. Candidate graph configurations are evaluated directly against the target objective, while gradients are used only where they are locally useful for refining shared continuous parameters. We evaluate EvoForest on structural break detection in the 2025 ADIA Lab Structural Break Challenge. EvoForest reaches 94.13\% ROC-AUC after 600 evolution steps, exceeding the publicly reported winning score of 90.14\% under the same evaluation protocol. Beyond the score itself, the experiment supports the broader claim of the paper: strong predictive performance can emerge from explicit search over computations, guided by diagnostics and refined by limited continuous optimization.

In this work, building on the EvoLattice internal-population representation \citep{yuksel2025evolattice}, we introduce a multi-alternative DAG for supervised learning that supports open-ended search over reusable computations, including callable higher-order transformations and persistent trainable parameters. Second, we present a hybrid optimization framework that separates structural evolution from local gradient-based refinement. Third, we introduce a diagnostic feedback loop in which model internals are transformed into explicit evolutionary guidance through structured reports and persistent memoranda. Fourth, we show that this search-first formulation yields parameter-efficient predictors optimized directly for non-differentiable cross-validation targets, while jointly discovering both domain-specific computations and representation-learning structure and parts of the downstream fitting rule itself through learned sample weighting and residual reweighting. More broadly, we argue that EvoForest exemplifies a wider shift in machine learning: from treating learning primarily as parameter fitting toward treating computation discovery as a first-class learning mechanism.

\section{Background and Related Work}

In many structured prediction problems, the main challenge is not fitting more parameters, but discovering the right computations: robust statistics, temporal summaries, invariances, interaction tests, nonlinear transforms, gates, or other domain-specific compositions. This is especially true when objectives are non-differentiable, only weakly aligned with standard surrogates, data are limited, or interpretability and domain priors matter. In such settings, scaling model size or hyper-tuning a fixed learner is often less effective than discovering better computations \citep{borisov2022deep,grinsztajn2022tree}. EvoForest addresses this setting by treating the learned object not simply as a weight tensor, but as an evolving computational structure. Automated feature engineering recognizes the importance of input transformations. Systems such as Deep Feature Synthesis and reinforcement-learning-based feature engineering generate composed transformations from predefined operator libraries \citep{kanter2015deep,khurana2018feature}. However, they are typically limited by fixed vocabularies, flat feature spaces, and one-shot scoring. Genetic programming goes further by explicitly searching over symbolic expressions \citep{koza1992genetic,espejo2010survey,lacava2019general}, but classical GP usually relies on tree-structured expressions with limited computation sharing and weak persistent memory. EvoForest builds on these ideas, but searches over reusable DAGs with shared intermediates, multiple alternatives per node, callable higher-order function families, and optional gradient-based refinement of persistent continuous parameters. Neural architecture search treats model structure as an optimization target \citep{zoph2017nas,liu2019darts,real2019regularized}, but still assumes that the main learned object is a neural network whose capacity lies primarily in trained weights. EvoForest is broader: it searches directly over computations, including statistics, transforms, similarity functions, gates, projections, and callable families. In this sense, it extends architecture search into \emph{computation search}. EvoForest is also related to random-feature methods, kernel methods, and mixture-of-experts models. Random-feature and kernel methods show that simple linear predictors can work well when paired with a rich nonlinear basis \citep{rahimi2007random,bach2017breaking,bartlett2002rademacher}. EvoForest adopts the same downstream principle, but replaces random or fixed bases with a discovered, task-adaptive computational basis. Its Ridge-based readout is therefore a deliberate choice: fast, parameter-efficient, and diagnostically transparent. Mixture-of-experts models likewise exploit modular computation through multiple specialists and sparse routing \citep{jacobs1991adaptive,shazeer2017moe,fedus2022switch}. EvoForest shares this modular intuition, but the modules are not fixed in advance: they are invented, revised, pruned, and recombined through search, and selection is performed not only by routing but also by configuration search and lightweight downstream scoring.

Recent work uses LLMs to propose feature transformations and predictive signals \citep{zhang2024dynamic,gong2024evolutionary,yuksel2025alphaquant}. EvoForest builds on this direction, but shifts from flat feature generation to graph-structured computation search. The LLM is not merely a feature generator; it acts as a search operator that responds to diagnostics, execution feedback, and accumulated search history. EvoForest also relates to LLM-guided program synthesis and discovery. Recent systems use LLMs to generate, repair, and evolve executable code \citep{chen2021codex,li2022repair,yao2023react,ellis2021dreamcoder}. Systems such as FunSearch, AlphaEvolve, and ShinkaEvolve demonstrate that LLM-guided evolutionary search can discover useful algorithms and programs \citep{romera2024funsearch,alphaevolve2024,shinkaevolve2024}. Contemporary autonomous research systems such as AutoResearch further show that LLM agents can iteratively modify training code, run experiments, and optimize empirical outcomes in closed research loops \citep{karpathy2026autoresearch}. However, most of these methods operate in an \emph{external-population, monolithic-evolution} regime, where each candidate is internally single-path, and mutations overwrite prior structure. By contrast, EvoForest builds on EvoLattice \citep{yuksel2025evolattice}, which introduced an \emph{internal-population, multi-alternative} DAG representation in which each valid path defines a distinct candidate while alternatives remain persistent and reusable within a shared graph. This enables combinatorial search, alternative-level statistics, and non-destructive local edits. EvoForest extends this representation from program and agent evolution to supervised learning by adding callable transformation families, persistent trainable parameters, configuration-based evaluation, a parameter-light Ridge readout, and diagnostic-guided search over both domain-specific computations and representation-learning structure. It is therefore best viewed as a search-first machine learning framework built on the EvoLattice representation.

\section{Methodology}

\begin{figure}[H]
    \centering
    \includegraphics[width=0.82\textwidth]{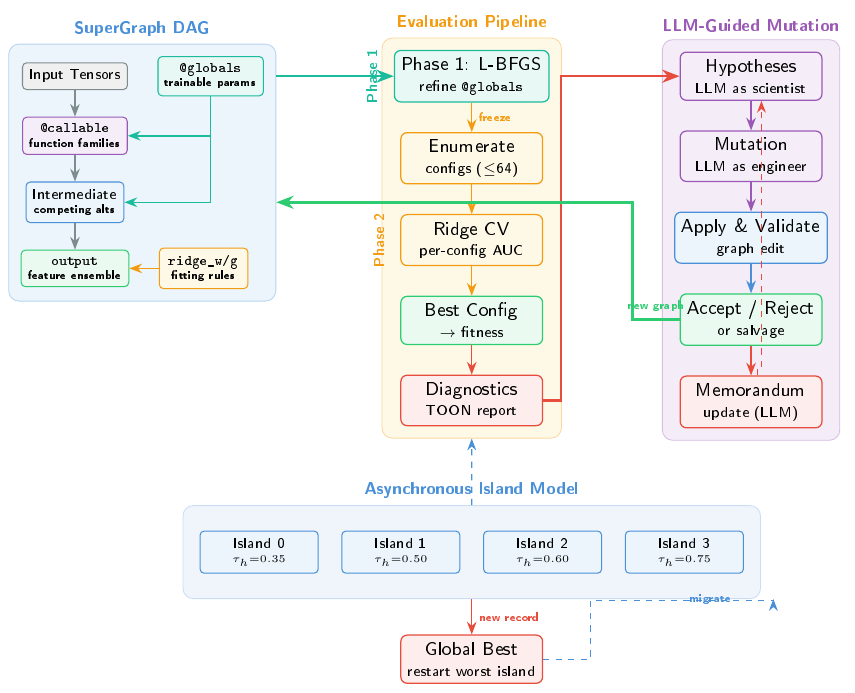}
    \vspace{-4pt}
    \caption{\textbf{EvoForest architecture.} EvoForest DAG (left), two-phase evaluation pipeline (center), LLM-guided scientist/engineer mutation loop (right), and asynchronous island model (bottom).}
    \label{fig:architecture}
\end{figure}

EvoForest is a hybrid neuro-symbolic system for feature discovery. An architectural overview is provided in Figure~\ref{fig:architecture} (Appendix~A). It represents a population of candidate computations inside a single directed acyclic graph, scores graph configurations with a lightweight linear model, extracts analytic diagnostics from that evaluator, and feeds those diagnostics back into an LLM-guided mutation loop. The framework couples five search processes: structural search over computational paths, function-family search over reusable callable transforms, feature search over output combinations, scorer search over sample weighting and residual reweighting, and continuous optimization over shared trainable parameters. Crucially, EvoForest does not search only over representations; it also searches over parts of the downstream estimation procedure itself. The central methodological idea is that EvoForest does not treat learning as optimization over a fixed parametric model. Instead, it treats learning as search over reusable computations supported by lightweight prediction and rich diagnostics. The LLM determines what computations should exist and how they should be recombined; gradient descent refines the continuous parts of those computations; the linear evaluator determines which discovered features matter; and the diagnostic loop turns that judgment into evidence for future search. This division of labor is the core reason EvoForest differs from flat feature generation, classical genetic programming, and conventional architecture search. EvoForest is a directed acyclic graph whose nodes contain persistent alternative implementations. These alternatives coexist inside the same graph and define an internal population of candidate computations. This representation separates which internal computations are selected from which output features are combined, allowing many features to reuse shared intermediates while still being judged jointly by the downstream evaluator. The DAG is represented with four entity types:

\begin{itemize}
    \item \textbf{Globals store.} The \texttt{@globals} store contains named trainable tensors shared across nodes and configurations. These parameters support learnable projections, gates, kernels, and activation families. The LLM may add new parameter entries, while gradient-based optimization refines their values.

    \item \textbf{Callable nodes.} Nodes prefixed with \texttt{@} other than \texttt{@globals} are higher-order callable nodes. Their alternatives return functions such as activations, projections, normalization rules, or interaction kernels. Downstream nodes consume these selected functions as arguments. Callable nodes therefore make reusable transformation families explicit search objects.

    \item \textbf{Intermediate nodes.} Intermediate-node alternatives are competing implementations that return tensors. A configuration selects one alternative per reachable intermediate node, thereby defining a concrete computational backbone.

    \item \textbf{Output node.} Output-node alternatives are candidate features rather than mutually exclusive choices. For a fixed configuration, all output alternatives are evaluated and stacked into a feature matrix that is scored jointly.

    \item \textbf{Fitting nodes.} Fitting-node alternatives alter the fitting dynamics of the downstream readout. The \texttt{ridge\_w} node produces per-sample weights, while the \texttt{ridge\_g} node defines a residual-to-weight mapping used inside iterative reweighted least squares.
    
\end{itemize}

A configuration selects one alternative for each reachable intermediate and callable node. Evaluating all output alternatives for a configuration yields a feature matrix over the dataset. In principle, the number of configurations grows multiplicatively with the number of multi-alternative nodes, so EvoForest caps the number of evaluated configurations in practice. The DAG representation enables computational sharing. If two configurations agree on a node and all of its ancestors, that subcomputation is identical and can be reused. EvoForest therefore caches intermediate values at the level of ancestor-conditioned subpaths rather than recomputing every feature independently. This is a major advantage over flat feature-generation systems: useful intermediate statistics, transforms, and projections can be computed once and reused across many downstream outputs. A major architectural feature of EvoForest is the callable node. These nodes lift reusable function families into first-class citizens of the search space. A callable node may contain alternatives such as a SiLU activation, a KAN-style spline activation, an RBF transform, a learnable gate, a linear projection, or a normalization rule. During evaluation, a configuration selects one callable alternative, and downstream nodes invoke the selected function on their inputs. This mechanism turns function-family choice into an explicit combinatorial axis of search. The system is therefore not limited to deciding which statistics to compute; it also searches over how those statistics should be transformed. The same intermediate signal can be routed through different activation families in different configurations, allowing the evaluator to determine which pairing of signal and transformation is most useful. The \texttt{@globals} store bridges symbolic search and continuous optimization. Its entries are persistent trainable tensors represented as \texttt{nn.Parameter} objects. The LLM controls the structure of this store by proposing new parameter entries, their shapes, and their initialization rules; gradient-based optimization controls the values of these parameters during evaluation. To preserve continuity across evolutionary steps, the parameter store follows append-only semantics at the structural level: new entries may be added, but existing trained entries are not overwritten by mutation. Unused entries are later pruned when they are no longer referenced by any reachable computation. This creates a form of parameter inheritance in which useful learned continuous components survive structural evolution. EvoForest not only searches over feature computations, callable transformations, and output ensembles. It also searches over sample reweighting policies through \texttt{ridge\_w} and robust residual weighting laws through \texttt{ridge\_g}. As a result, the downstream fitter is no longer a fixed evaluator applied after representation discovery; it becomes part of the search space itself. This allows the system to adapt not only the representation being fit, but also the fitting dynamics used to evaluate and refine that representation, akin to curriculum learning and robust objective shaping.

EvoForest uses an asynchronous island model to sustain diversity. Multiple islands evolve in parallel, each maintaining its own graph, diagnostics, and memorandum. When one island achieves a new global best, its graph and learned parameters can be transferred to a weaker island, which then continues searching from that stronger state under its own future mutations. This design combines independent exploration with periodic knowledge transfer. Some islands make coarse exploratory jumps; others accumulate incremental refinements. Migration allows successful structure and trained continuous components to spread without collapsing the population into a single search trajectory. Each island maintains a persistent experiment memorandum (Appendix~\ref{sec:memorandum}): a short LLM-authored document that compresses recent diagnostics, score changes, mutation outcomes, and runtime failures into a compact search summary. The memorandum records which structures are dominating, which redundancy clusters should be pruned, where nonlinear opportunities remain, which mutation types have recently helped, and which implementation patterns have caused errors. This interpreted memory is an important part of the method. Without it, the LLM would need to rediscover search context from raw diagnostics at every step. With it, mutation can be conditioned not only on the current graph snapshot but also on the trajectory of what has been working and failing. Evolution proceeds through two LLM stages (Appendix~\ref{sec:prompts}) that separate speculative search from implementation discipline, reducing the chance that every creative idea becomes an expensive, invalid graph edit. First, the LLM acts as a \emph{scientist} and proposes a small set of candidate structural hypotheses conditioned on the current graph, diagnostics, memorandum, task summary, and recent feedback. These may involve pruning redundant outputs, adding alternatives to productive bottlenecks, introducing new callable families, composing existing signals more deeply, or adding trainable parameters to parts of the graph that show residual nonlinear structure. Then, the LLM acts as an \emph{engineer} and evaluates the proposed hypotheses, discards weak or inconsistent ones, and emits concrete graph mutations in structured form. These mutations may add or remove node alternatives, introduce new callable nodes, append new \texttt{@globals} entries, or simplify dead or redundant substructures. To prevent unbounded growth and maintain validity, EvoForest performs graph maintenance. Runtime-failed alternatives are pruned; unreachable subgraphs are removed; empty nodes are deleted; duplicate implementations are collapsed; trivial identity chains are inlined; and unused global parameters are discarded when safe to do so. Output features with persistently zero contribution may also be pruned, while productive bottleneck nodes may be preserved or expanded. This maintenance step enables a recurring behavior observed in evolution: the graph does not simply grow monotonically, but alternates between expansion, consolidation, and internal diversification.

\noindent\textbf{Phase 1: parameter refinement.}
If the current graph uses trainable globals, EvoForest first optimizes those parameters on a fixed structural path using L-BFGS together with an ephemeral linear readout layer. A short gradient probe determines whether any \texttt{@globals} entry actually influences the selected path; if not, this phase is skipped. This refinement stage lets the system tune continuous elements (knot positions, projection matrices, kernel centers, or gate scales) before structural scoring.

\noindent\textbf{Phase 2: structural scoring.}
The refined parameters are then frozen, caching is enabled, and all candidate configurations are evaluated with a Ridge model under cross-validation. This separation is deliberate: gradient-based refinement improves the quality of the current computational basis, while the downstream linear evaluator provides stable, interpretable fitness signals for structural search.

For each configuration, EvoForest evaluates the output feature matrix and scores it using stratified cross-validated Ridge regression, optionally augmented with configuration-dependent sample weights from \texttt{ridge\_w} and residual-based reweighting through \texttt{ridge\_g}. Methodologically, this means EvoForest searches not only over feature spaces but also over fitting rules. The system can therefore discover task-dependent forms of sample emphasis, de-emphasis, and robustness, rather than relying on a single fixed downstream estimator. In effect, EvoForest learns aspects of both representation and estimation. The feature matrix is standardized, the Ridge solution is computed in closed form from a batched singular value decomposition, and the regularization strength is selected analytically from a small log-scale grid using a leave-one-out error criterion. The graph-level fitness is the best configuration score achieved by the current graph. This configuration-based evaluation is critical. It separates intermediate-node selection, which defines a structural hypothesis, from output-feature combination, which is handled jointly by the downstream linear model. As a result, EvoForest does not evaluate features one by one in isolation; it evaluates reusable computational backbones together with the feature sets they support. Ridge regression is chosen for two reasons: it is computationally light enough to sit inside an evolutionary loop, and its closed-form solution exposes rich interpretable diagnostics that can be computed analytically and fed back into search. Every alternative in the graph accumulates statistics over the configurations and features in which it participates, including summary quality estimates and age. These statistics provide a compressed local history of which components have been broadly useful versus narrowly situational. After mutation, the candidate graph is evaluated. If its score matches or exceeds the current graph, it is accepted. Otherwise, EvoForest may still salvage locally beneficial alternatives from the rejected candidate into the incumbent graph. This salvage mechanism is important in a structured graph setting: an overall mutation can fail while still containing a valuable local component. After configuration scoring, EvoForest fits a global Ridge model over the valid feature pool and extracts a rich diagnostic profile for each output feature. These diagnostics include normalized importance, coefficient sign, standalone predictive strength, redundancy with other features, rank-based target association, class-separation effect size, exact linear SHAP decompositions, residual-feature correlations, residual quadratic correlations, and cross-fold weight stability. Additional node-level summaries are obtained by aggregating the diagnostics of all outputs that depend on a given intermediate or callable alternative. The purpose of these diagnostics is not merely analysis after the fact. They are the primary evidence used to steer search. A feature that is highly predictive alone but redundant in the ensemble suggests one kind of mutation; a feature with strong residual correlation suggests another; a node whose descendants contribute unstable weights suggests yet another. EvoForest converts the internals of a simple linear model into structured search guidance, which summarizes global graph context, per-feature diagnostics, and per-node or per-alternative aggregates in a dense machine-readable tabular form (Appendix~\ref{sec:toon}) that allows the LLM to reason not only about which features are important, but also about why they are, where redundancy is accumulating, which nonlinearities are unmodeled, and substructures that appear productive or brittle.

\section{Experiments}
\label{sec:experiments}

We evaluate EvoForest on the \textbf{ADIA Lab Structural Break Detection Challenge}~\citep{crunchdao2025structuralbreak_leaderboard}, a recent high-profile machine-learning competition for binary structural-break detection in univariate time series with a \textbf{USD 100,000 prize pool}. Each example provides a full sequence together with a designated boundary point, and the goal is to output a score in $[0,1]$ representing the likelihood that a genuine structural break occurred at that boundary. Because this is a live, high-stakes competition task, we treat the public leaderboard as the most meaningful external benchmark rather than attempting partial reimplementations of heterogeneous agentic systems under unmatched settings or diluting the evaluation across easier, less competitive problems. In particular, industry-developed state-of-the-art method, AIDE, is an autonomous, LLM-based machine-learning competition agent based on iterative solution generation, execution, evaluation, selection, and refinement through \emph{Solution Space Tree Search}~\citep{schmidt2024aide}; \emph{MLE-bench} likewise describes AIDE as a Kaggle-purpose-built tree-search scaffold rather than a directly comparable structural-search learner~\citep{jun2025mlebench}. We therefore use the open leaderboard as the primary reference point and leave direct head-to-head benchmarking against similar agentic competition systems for future work. Under this framing, EvoForest's final score of \textbf{94.13} AUC exceeds the publicly listed winning score of \textbf{90.14}. 

The training set contains 10{,}001 labeled series (2{,}909 positives, 7{,}092 negatives). Sequence lengths range from 1258 to 3478 (mean 2371); pre-break lengths range from 1000 to 2499 and post-break from 250 to 999. The raw values are highly heterogeneous, making the task difficult due to variable scale, variable segment length, and subtle regime shifts. Candidate outputs are evaluated with the challenge-aligned scorer: stratified 3-fold cross-validation with a Ridge model, per-fold alpha selection via leave-one-out MSE, and final score given by mean ROC--AUC across folds. EvoForest also generates a \emph{task context summary} for domain adaptation. At startup, the LLM reads the task source code and dataset statistics and compresses the tensor inventory, scorer mechanics, and implementation constraints into a cached summary injected into subsequent hypothesis-generation calls so that mutations remain grounded in the actual task interface. The model receives pre-segmented tensors for the full sequence, pre-break segment, post-break segment, their masks, and scalar metadata. The full sequence is padded to length 3478, the pre-segment to 2499, and the post-segment to 999.

\begin{figure}[H]
    \centering
    \begin{subfigure}[t]{0.48\textwidth}
        \centering
        \includegraphics[width=\linewidth]{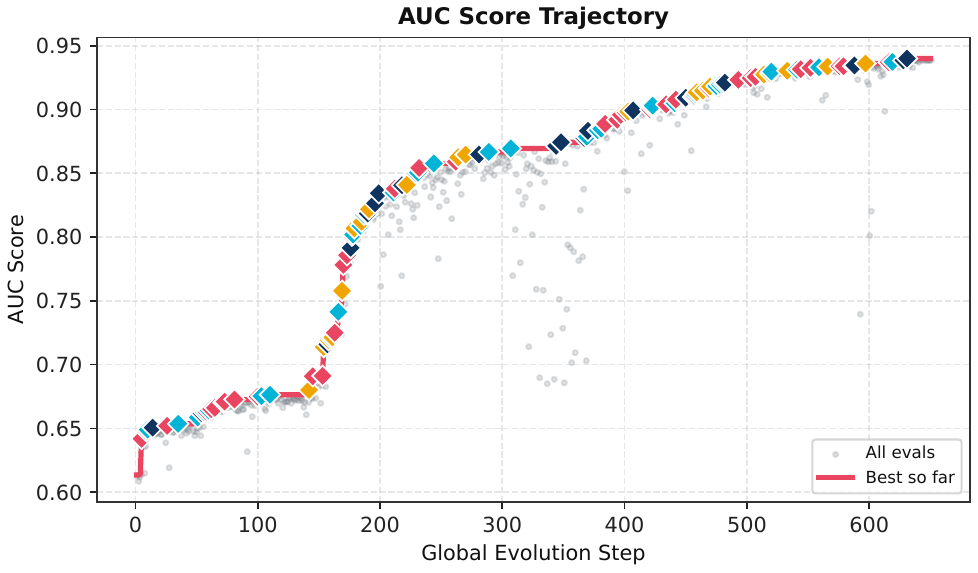}
        \caption{}
    \end{subfigure}
    \hfill
    \begin{subfigure}[t]{0.48\textwidth}
        \centering
        \includegraphics[width=\linewidth]{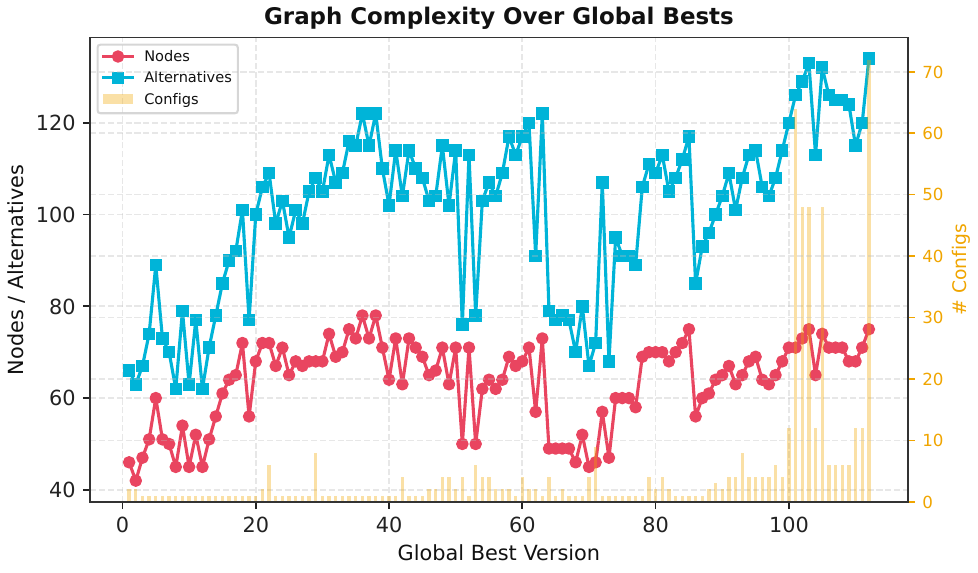}
        \caption{}
    \end{subfigure}\\[-2pt]
    \begin{subfigure}[t]{0.48\textwidth}
        \centering
        \includegraphics[width=\linewidth]{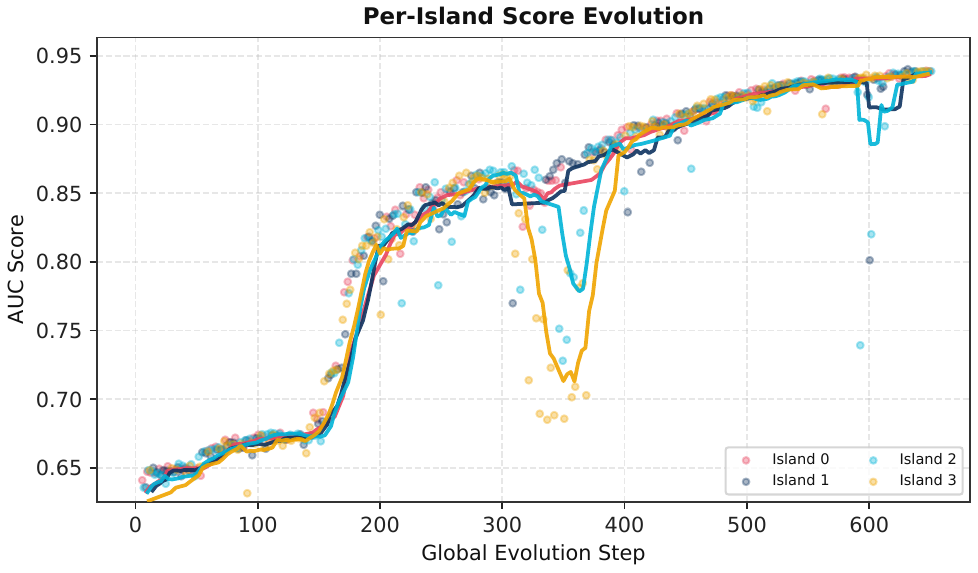}
        \caption{}
    \end{subfigure}
    \hfill
    \begin{subfigure}[t]{0.48\textwidth}
        \centering
        \includegraphics[width=\linewidth]{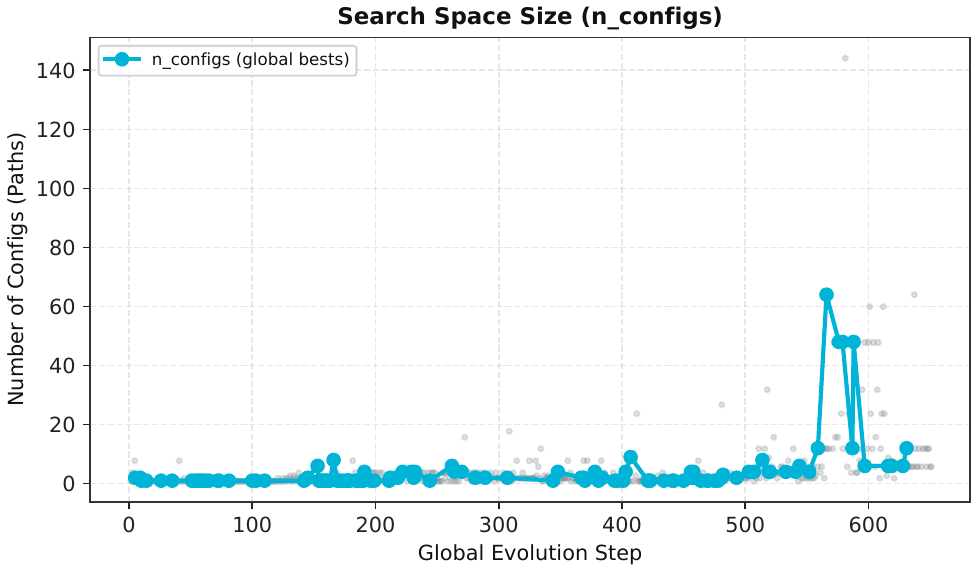}
        \caption{}
    \end{subfigure}
    \vspace{-4pt}
    \caption{\textbf{Optimization and search-space diagnostics.}
    (a) Best-so-far ROC--AUC. (b) Graph complexity over time. (c) Per-island scores. (d) Configuration count.}
    \label{fig:main_results}
\end{figure}

\begin{figure}[H]
    \centering
    \begin{subfigure}[t]{0.48\textwidth}
        \centering
        \includegraphics[width=\linewidth]{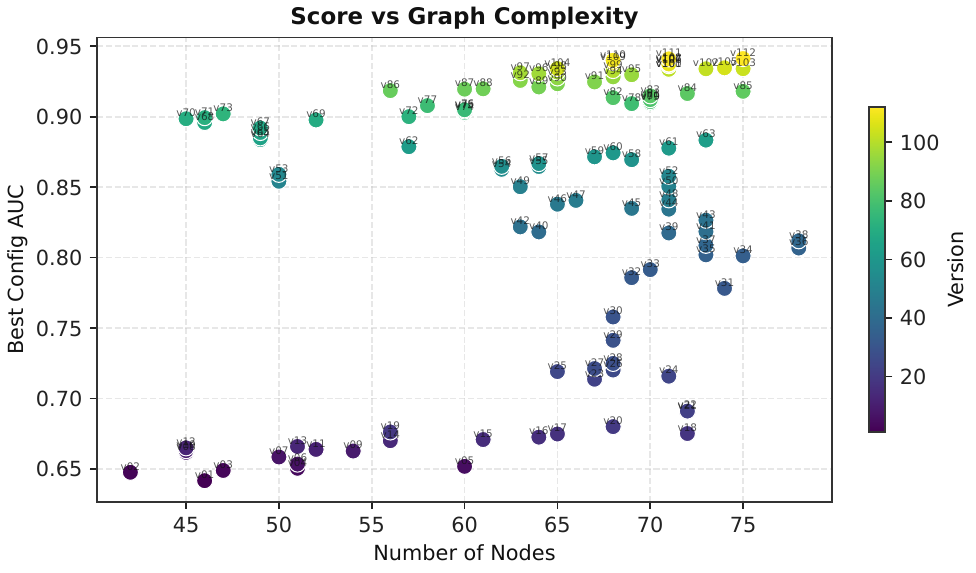}
        \caption{}
    \end{subfigure}
    \hfill
    \begin{subfigure}[t]{0.48\textwidth}
        \centering
        \includegraphics[width=\linewidth]{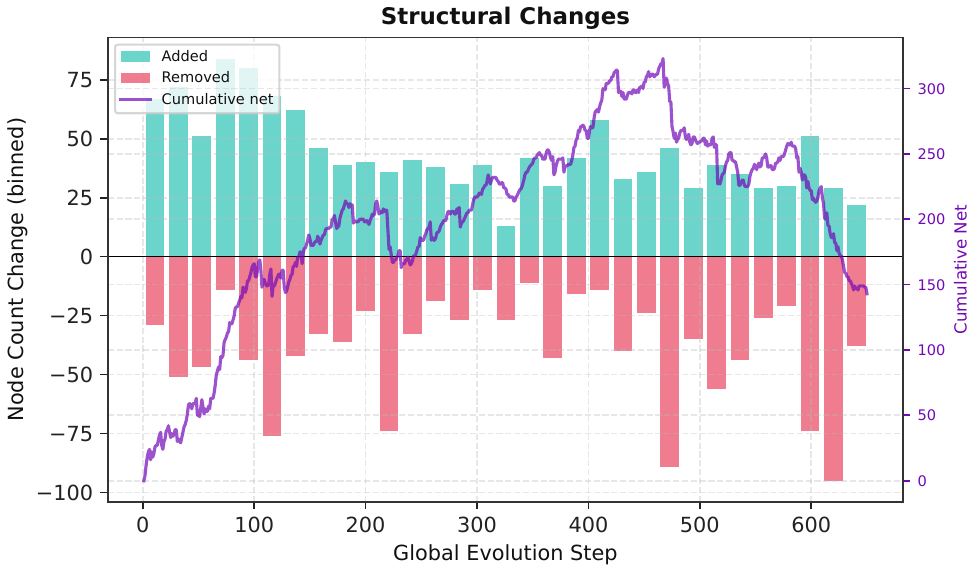}
        \caption{}
    \end{subfigure}\\[-2pt]
    \begin{subfigure}[t]{0.48\textwidth}
        \centering
        \includegraphics[width=\linewidth]{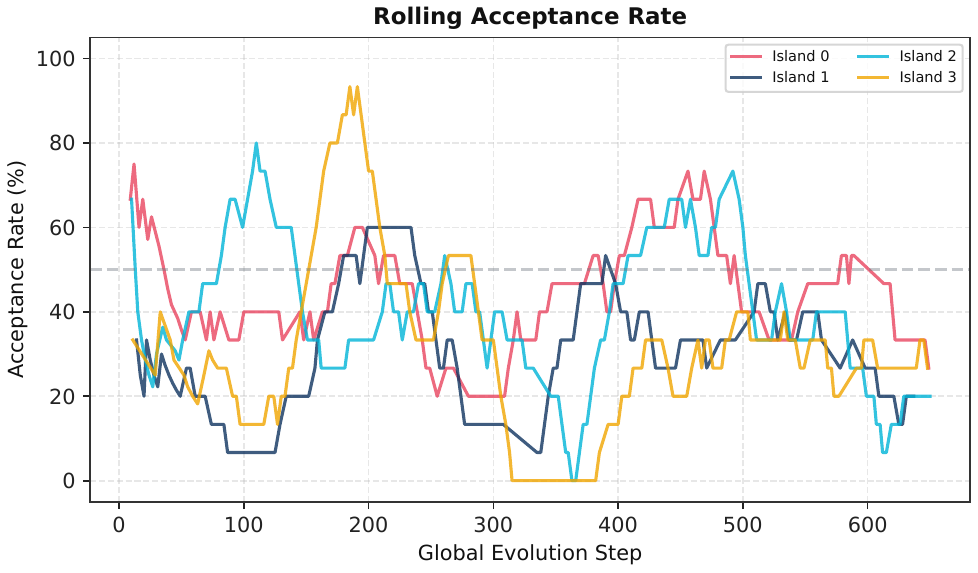}
        \caption{}
    \end{subfigure}
    \hfill
    \begin{subfigure}[t]{0.48\textwidth}
        \centering
        \includegraphics[width=\linewidth]{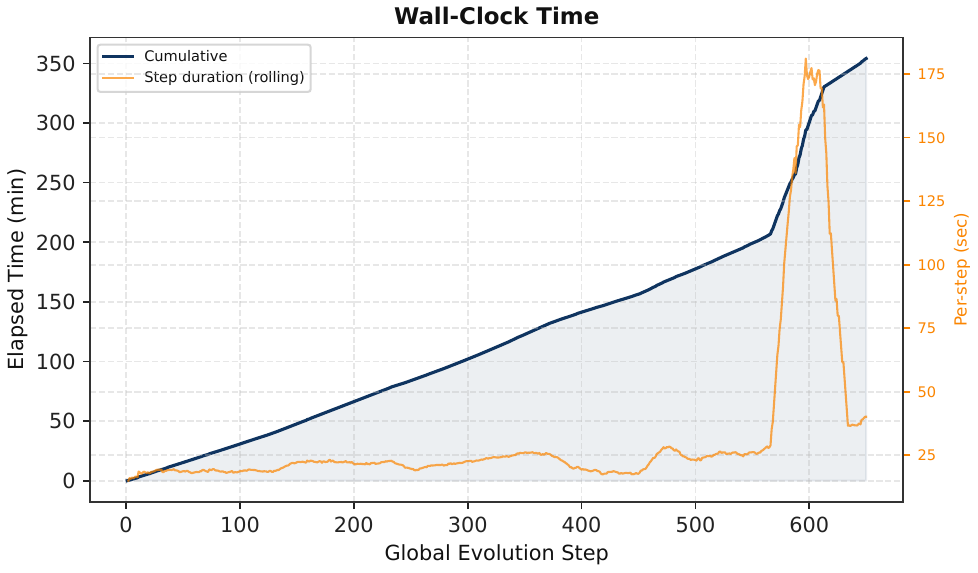}
        \caption{}
    \end{subfigure}
    \vspace{-4pt}
    \caption{\textbf{Structural and systems diagnostics.}
    (a) Score vs.\ complexity. (b) Structural changes. (c) Acceptance rates. (d) Wall-clock time.}
    \label{fig:main_systems}
\end{figure}

\begin{figure}[H]
    \centering
    \begin{subfigure}[t]{0.48\textwidth}
        \centering
        \includegraphics[width=\linewidth]{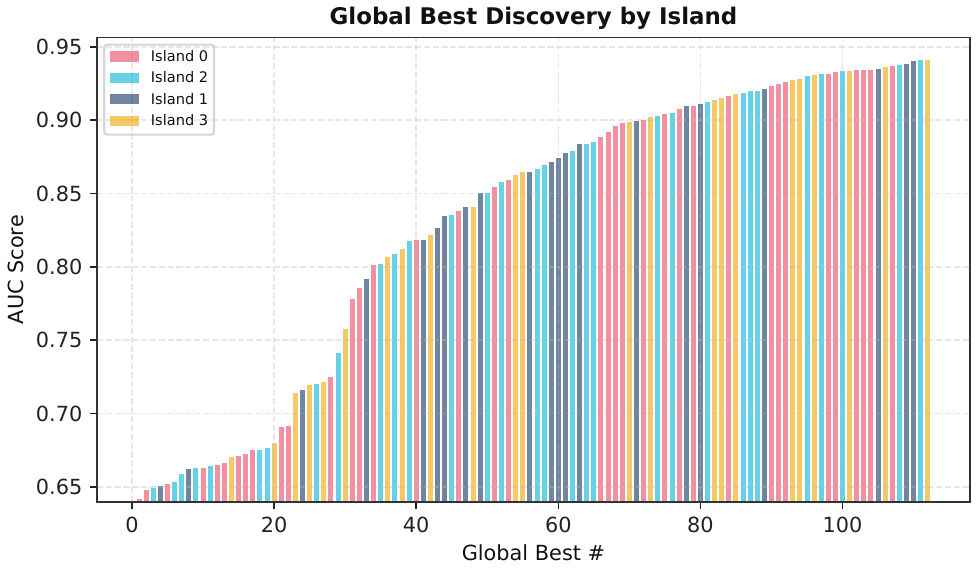}
        \caption{}
    \end{subfigure}
    \hfill
    \begin{subfigure}[t]{0.48\textwidth}
        \centering
        \includegraphics[width=\linewidth]{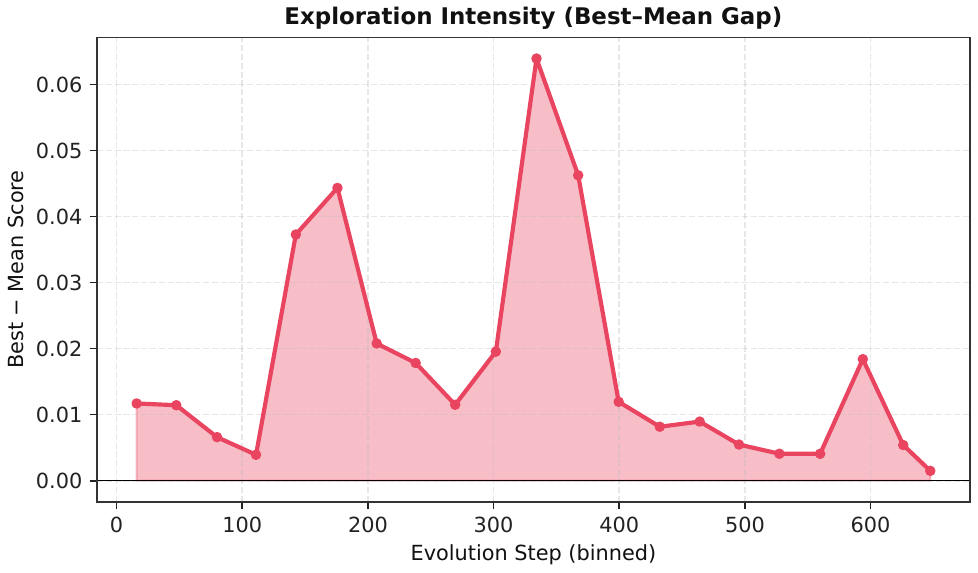}
        \caption{}
    \end{subfigure}\\[-2pt]
    \begin{subfigure}[t]{0.48\textwidth}
        \centering
        \includegraphics[width=\linewidth]{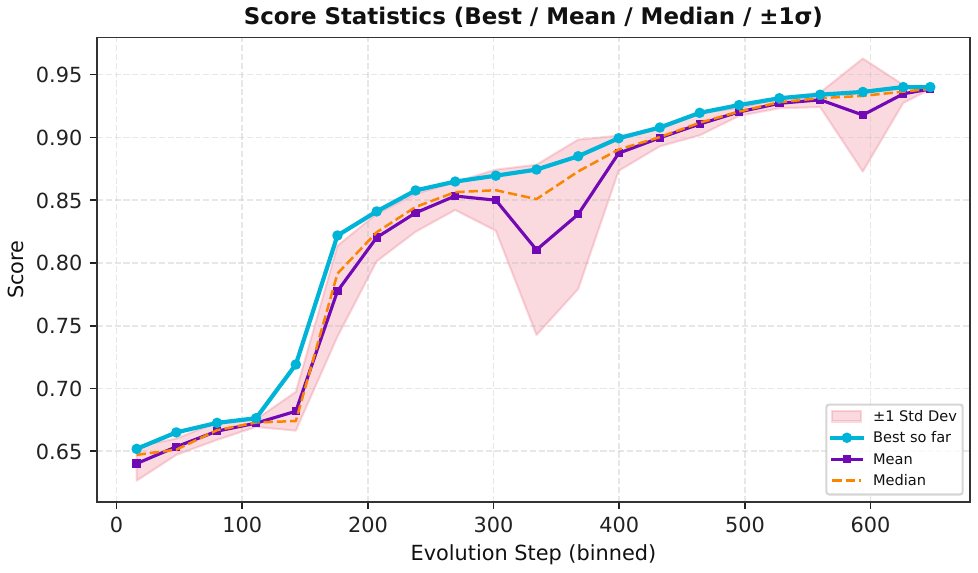}
        \caption{}
    \end{subfigure}
    \hfill
    \begin{subfigure}[t]{0.48\textwidth}
        \centering
        \includegraphics[width=\linewidth]{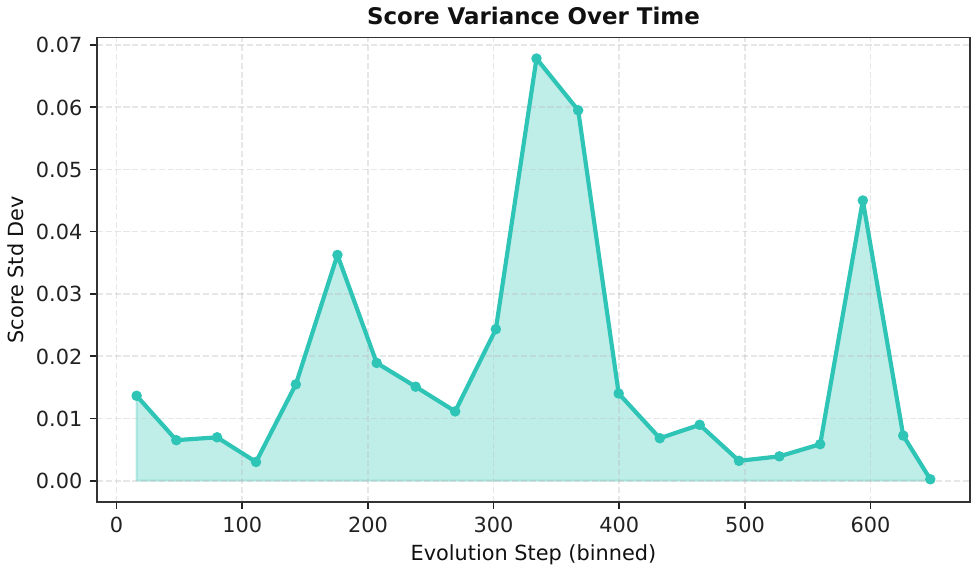}
        \caption{}
    \end{subfigure}
    \vspace{-4pt}
    \caption{\textbf{Additional diagnostics.}
    (a) Discoveries by island. (b) Best--mean gap. (c) Score statistics. (d) Score variance.}
    \label{fig:app_diagnostics_1}
\end{figure}

Table~\ref{tab:adia_context} (Appendix~B) provides a benchmark comparison with the public leaderboard entries.
Our long-horizon experiment uses \textbf{4 asynchronous islands}, each on a dedicated GPU. Phase-1 hypothesis generation uses fixed island-specific temperatures $\{0.35, 0.5, 0.6, 0.75\}$, while Phase-2 mutation synthesis uses zero temperature. The evaluator caps the number of enumerated intermediate configurations per evaluation at \textbf{64}. Persistent parameters in \texttt{@globals} are optimized with L-BFGS, using \textbf{200} iterations for the initial cold-start evaluation and \textbf{20} iterations thereafter. The experiment spans a \textbf{600-step} asynchronous run with \textbf{112} global-best versions, or roughly one new global record every five to six steps. The overall mutation acceptance rate is approximately \textbf{33.5\%}. The final best cross-validated ROC--AUC is \textbf{0.9413}, achieved by global best \textbf{v112} at evolution step \textbf{600}, attributed to \textbf{Island~3}. Figures~\ref{fig:app_diagnostics_1}(a) and \ref{fig:main_results}(c) show that all four islands contribute meaningfully to the global-best frontier. Improvements are interleaved across islands rather than dominated by a single worker, supporting the intended division of labor: cooler islands refine promising motifs, while warmer islands inject more structural novelty. The rolling acceptance rates in Figure~\ref{fig:main_systems}(c) show alternating periods of aggressive exploration and conservative consolidation. Figure~\ref{fig:main_systems}(d) reports cumulative wall-clock time together with rolling step duration. Runtime grows approximately linearly overall, but with visible late-stage spikes in per-step cost. These spikes align with the high-configuration regime shown in Figure~\ref{fig:main_results}(d). Thus, richer internal branching improves representational power but also increases evaluation cost.

Figure~\ref{fig:main_results}(a) shows the global score trajectory. The best-so-far curve exhibits a staircase profile: extended periods of gradual progress are interleaved with sharper jumps, consistent with a search process in which many accepted mutations are local refinements while occasional structural innovations unlock qualitatively stronger feature bases. The run progresses through three broad phases that align with the milestone table. The earliest milestones deliver steady but relatively small gains, establishing the initial jump-sensitive and distribution-aware feature basis. This is followed by a more disruptive middle period, where several larger improvements appear in close succession as the system introduces richer nonlinear interactions, spectral-statistical couplings, and more structured multi-scale compositions. After this rapid ascent, the search transitions into a longer refinement phase in which improvements become smaller but remain persistent. This late stage is especially important: although each gain is modest, the system continues to improve through recombination, pruning, and stabilization of previously discovered motifs, ultimately reaching \textbf{0.9413}. Overall, the trajectory is consistent with a search process that first discovers useful computational building blocks and then consolidates them into a stronger and more stable graph, as supported by distributional summaries in Figures~\ref{fig:app_diagnostics_1}(c) and \ref{fig:app_diagnostics_1}(d). The score rises steadily over time, while the spread of candidate scores first widens during the disruptive middle phase and then contracts during late consolidation, indicating that breakthroughs appear first at the frontier before diffusing through later search.

Performance gains are accompanied by sustained structural growth. Figure~\ref{fig:main_results}(b) tracks graph complexity over global-best versions. By the final best, EvoForest reaches \textbf{75} nodes, \textbf{53} output features, \textbf{134} alternatives across intermediate nodes, \textbf{72} joint intermediate configurations, and maximum DAG depth \textbf{7}. Compared with early global-best graphs, this reflects a substantial broadening of the representational search space. Importantly, complexity growth is not pure accumulation. Figure~\ref{fig:main_systems}(b) shows alternating periods of expansion and pruning. Early in the run, net additions dominate as the system discovers useful statistical, spectral, and interaction motifs. Later, removals become more common, suggesting that once stronger motifs are available, the system can discard redundant structure while maintaining or improving performance. Figure~\ref{fig:main_systems}(a) further shows that better performance often correlates with greater complexity, but not monotonically. Some smaller graphs outperform earlier larger graphs, indicating that EvoForest is discovering better structure rather than merely more structure. More broadly, the graph behaves neither like monotonic network growth nor like pure pruning; it cycles through expansion, simplification, and reorganization. One of the central design goals of EvoForest is to encode a population of candidate computations inside a single graph through node-local alternatives and joint configurations. Figure~\ref{fig:main_results}(d) shows how the effective number of configurations evolves. For much of the run, global-best graphs have relatively few active configurations, but the later stages enter a clearly higher-combinatorial regime. These late gains coincide not only with higher scores but also with richer internal branching, showing that the method benefits from preserving multiple competing computational pathways in the same DAG rather than collapsing prematurely to a single rigid pipeline. At the same time, the strongest graphs are not simply the deepest or most expanded: temporary growth in configuration space appears to help discover stronger structures, which are then consolidated into a more compact final graph.

Overall, the experiment provides strong evidence for the search-first view advocated in this paper. The final best graph achieved $0.9413$ AUC after 600 evolution steps, ending with 75 nodes, 134 alternatives, 53 features, and depth~7. The search process was \emph{productive}: it improved the seeded system by nearly 0.30~AUC while continuing to discover new global bests throughout a long run. It was also \emph{structural rather than merely parametric}: large improvements coincide with changes in nodes, alternatives, and configurations, while depth remains nearly constant. The key axis of learning is the discovery and reuse of better computations, not simply deeper composition. The island model was \emph{functionally important}: hotter islands contributed large jumps while cooler islands consolidated strong motifs. Better performance does \emph{not} require monotonically increasing complexity; several strong jumps were followed by simplification, and the final graph is not the largest encountered. This supports the core design of maintaining multi-alternative reusable structure while allowing aggressive pruning and consolidation. Table~\ref{tab:milestones} (Appendix~E) summarizes selected milestones from the 600-step run, showing that the most important gains arise from structural innovation rather than numeric tuning. More broadly, the results suggest that feature quality matters more than raw dimensionality, and that late-stage gains come from better organization, selection, and reuse within an already strong computational basis, though at increased computational cost.

\section{Conclusion}

We introduced EvoForest, a hybrid neuro-symbolic framework that treats supervised learning as search over reusable computations rather than optimization within a fixed model family. It evolves a multi-alternative DAG of competing computations, callable transformation families, output features, and persistent trainable parameters, while a lightweight Ridge-based evaluator provides both fitness and diagnostic feedback. On the ADIA Lab Structural Break Detection Challenge, EvoForest reached \textbf{94.13\% ROC--AUC}, exceeding the publicly reported winning score under the same evaluation protocol. The results show that the main gains come from discovering, recombining, and refining useful computational motifs, while the multi-alternative graph and asynchronous island evolution help sustain diversity and improvement over long horizons. More broadly, EvoForest supports a search-first view of machine learning: in many structured problems, the central challenge is not only learning weights, but discovering the right computations. Notably, the computations discovered by EvoForest can yield transferable algorithmic knowledge: analyzing the evolved graphs revealed a recurring multi-space feature allocation motif that was distilled into Multi-Space Random Features (MSRF), a competitive standalone time-series method that outperforms MiniRocket on the 112-dataset UCR archive using 86\% fewer features (Appendix~\ref{app:msrf}). Our results suggest that explicit computation discovery can be a practical and powerful complement to parameter-centric learning.

\bibliography{references}

\appendix

\section{Architecture Overview}

\vspace{-8pt}
\section{Benchmark Comparison}
\vspace{-6pt}

\begin{table}[H]
\centering
\caption{Benchmark comparison for the ADIA structural-break task. The public leaderboard provides the most direct external benchmark for this live competition. We also include the \texttt{Weco AI} leaderboard entry as a baseline because they publicly introduced \textsc{AIDE}, an autonomous competition agent; independent evaluation in \emph{MLE-bench} characterizes AIDE as a Kaggle-specialized tree-search scaffold.}
\label{tab:adia_context}
\small
\begin{tabular}{p{1.8cm}p{2.0cm}p{6.4cm}c}
\toprule
Entry / system & Category & Description / relevance & Score \\
\midrule
EvoForest (ours) & Search-first & Multi-alternative DAG with reusable computations, callable families, trainable globals, configuration search, and diagnostic-guided LLM mutation. & \textbf{94.13} \\
alphabot & Challenge winner & Top public entry on the final leaderboard; strongest direct reference. & 90.14 \\
homely-rat & Industry entry & Weco AI using \textsc{AIDE}, an autonomous ML competition agent. & 78.32 \\
\bottomrule
\end{tabular}
\end{table}

\clearpage
\section{Experimental Figures}
\vspace{-4pt}

\vspace{-2pt}

\clearpage

\vspace{-2pt}
\section{Evolved Graph Excerpt}
\vspace{-4pt}
\begin{evolisting}
'@globals':
  rp_W: torch.randn(8, 4) * (1.0 / 8.0 ** 0.5)
  rp_b: torch.zeros(4)
  mr_base_w: torch.randn(16, 9) * (1.0 / 9.0 ** 0.5)
  mr_kmask: torch.ones(16, 1, 9)
  mr_dil: torch.tensor([1,1,1,1,2,2,2,2,4,4,4,4,8,8,8,8])
  rff_W: torch.randn(256, 8) * 0.1
  rff_b: torch.randn(256) * 0.1
'@mr_ppv':
- 'lambda seg, mask, globals: torch.cat([((F.conv1d(
  seg.unsqueeze(1), globals["mr_base_w"][i:i+1].unsqueeze(1),
  padding=int(globals["mr_dil"][i].item())*4,
  dilation=int(globals["mr_dil"][i].item()))>0).float()
  .sum(2) / F.conv1d(mask.float().unsqueeze(1),
  globals["mr_kmask"][i:i+1],
  padding=int(globals["mr_dil"][i].item())*4,
  dilation=int(globals["mr_dil"][i].item())).sum(2)
  .clamp(min=1.0)) for i in range(16)], dim=1)'
output:
- '... (pre_mrocket_ppv - post_mrocket_ppv).abs()'
- '... adaptive_cusum'
ridge_g:
- 'lambda r: torch.where(r.abs()<=1.0, torch.ones_like(r),
  1.0/r.abs().clamp(min=1e-6, max=1e6))'
\end{evolisting}

\footnotesize
\noindent A minimal excerpt illustrating the main entity types: trainable tensors (\texttt{@globals}), callable transforms (\texttt{@mr\_ppv}), intermediate nodes, output features, and fitting rules.
\normalsize

\clearpage
\section{Global-Best Milestones}

\footnotesize
\setlength{\LTleft}{0pt}
\setlength{\LTright}{0pt}

\begin{longtable}{lllllp{8.6cm}}
\caption{Global-best milestones for the long-horizon ADIA run.}
\label{tab:milestones} \\
\toprule
Version & Step & Island & AUC & $\Delta$AUC & Phase / key innovation \\
\midrule
\endfirsthead
\toprule
Version & Step & Island & AUC & $\Delta$AUC & Phase / key innovation \\
\midrule
\endhead
\bottomrule
\endfoot
v001 & 1   & 0-3 & 0.6120 & ---     & \textbf{Bootstrap.} LLM generated seed; jump/local-MAD ratio and variance-jump interaction establish the initial pre/post contrast basis. \\
v005 & 8   & 1 & 0.6418 & +0.0298 & \textbf{Bootstrap.} First strong jump from skewness/kurtosis contrasts, jump-absolute features, and a soft-gated nonlinear pathway. \\
v009 & 18  & 2 & 0.6506 & +0.0088 & \textbf{Bootstrap.} Spectral-entropy pairing, FFT-slope fusion, and mean--std interaction improve coarse segment discrimination. \\
v014 & 42  & 0 & 0.6640 & +0.0134 & \textbf{Early feature expansion.} Spectral entropy difference, Gaussian KL divergence, and an early MLP-style gate enrich the feature family. \\
v018 & 74  & 2 & 0.6726 & +0.0086 & \textbf{Early feature expansion.} Multi-scale windowed variance ratio, refined pruning, and CUSUM-conditioned gating improve robustness. \\
v022 & 111 & 2 & 0.6789 & +0.0063 & \textbf{Early feature expansion.} Joint gating, mixed RBF-style similarity, and higher-order moment contrasts expand the candidate family. \\
v026 & 145 & 0 & 0.6911 & +0.0122 & \textbf{Pre-transition.} Length-gated compositions, FFT--variance fusion, and richer output alternatives prepare the steep improvement phase. \\
v027 & 154 & 3 & 0.7137 & +0.0226 & \textbf{Transition jump.} A second cross-projection attention head and parallel length gate trigger the first major structural jump. \\
v030 & 166 & 2 & 0.7414 & +0.0277 & \textbf{Transition jump.} Competing length gates, cross-projection attention variants, and callable diversification sharply increase discriminative capacity. \\
v031 & 170 & 0 & 0.7782 & +0.0368 & \textbf{Transition jump.} MMD-style kernel scoring, multi-scale pyramids, and learned segment projections produce the steepest staircase in the run. \\
v034 & 176 & 3 & 0.7928 & +0.0146 & \textbf{Transition jump.} Structural simplification drops weak crossover paths while preserving stronger gated and attention-based branches. \\
v038 & 185 & 3 & 0.8120 & +0.0192 & \textbf{Transition jump.} Removal of weak self-attention cosine/pool substructures and stronger wavelet-band emphasis yield a cleaner backbone. \\
v044 & 208 & 0 & 0.8287 & +0.0167 & \textbf{Post-jump refinement.} Robust variance alternatives, spectral centroid differences, and callable-level gating improve the first refinement stage. \\
v049 & 228 & 1 & 0.8415 & +0.0128 & \textbf{Post-jump refinement.} RBF activation alternatives, mixed gating, and improved feature redistribution stabilize the graph in the low-0.84 regime. \\
v053 & 245 & 0 & 0.8554 & +0.0139 & \textbf{Post-jump refinement.} Spectral-band gates and FFT--variance alternatives with length-gated outputs open the next upward phase. \\
v058 & 266 & 2 & 0.8638 & +0.0084 & \textbf{Post-jump refinement.} Wavelet-energy plus robust variance alternatives refine the candidate set; gains become smaller and more incremental. \\
v061 & 305 & 0 & 0.8702 & +0.0064 & \textbf{Mid-run consolidation.} Output-family consolidation, improved pre/post contrasts, and selective pruning produce a broad plateau around 0.87. \\
v066 & 352 & 0 & 0.8749 & +0.0047 & \textbf{Mid-run consolidation.} Dynamic AR-style gating, third-order differences, and redistributed importance maintain a slow consolidation phase. \\
v071 & 389 & 3 & 0.8896 & +0.0147 & \textbf{Late re-acceleration.} Adaptive CUSUM gating and learned pre/post spectral filters restart progress after the plateau. \\
v076 & 412 & 1 & 0.9010 & +0.0114 & \textbf{Late re-acceleration.} Dedicated wavelet nodes, stronger length gating, and better cross-feature composition push the run through 0.90. \\
v083 & 446 & 3 & 0.9128 & +0.0118 & \textbf{High-AUC search.} Competing spectral, variance-ratio, and gating families create multiple strong alternatives in parallel. \\
v091 & 483 & 0 & 0.9226 & +0.0098 & \textbf{High-AUC search.} Improved callable reuse and a cleaner rich-output mix deliver steady late-run gains with less architectural churn. \\
v099 & 512 & 3 & 0.9308 & +0.0082 & \textbf{High-AUC search.} Consolidation favors the best length, spectral, and variance-ratio branches while removing unstable fusion paths. \\
v105 & 541 & 1 & 0.9361 & +0.0053 & \textbf{Final consolidation.} Global variance and pre/post std-ratio emphasis sharpen the downstream feature set in the low-0.93 regime. \\
v110 & 581 & 2 & 0.9400 & +0.0039 & \textbf{Final consolidation.} Contrastive length gate, meta-weighted FFT pathway, and mean-ratio family produce the last major improvement before saturation. \\
v112 & 600 & 3 & \textbf{0.9413} & +0.0013 & \textbf{Final consolidation.} Rich output alternatives, stable length-gate mixtures, and selective pruning of weaker crossover compositions yield the final best. \\
\end{longtable}
\normalsize
\noindent Table~\ref{tab:milestones} shows that large improvements are concentrated in a small number of structural jumps. Later gains become smaller but remain consistent, indicating a long consolidation phase rather than early saturation. The strongest improvements are distributed across multiple islands, supporting the intended exploration--exploitation division of labor.

\section{Prompt Templates}
\label{sec:prompts}

The mutation loop uses a two-phase LLM interaction: a \emph{scientist} generates hypotheses from diagnostics and memorandum; an \emph{engineer} translates them into YAML mutations. Both operate under shared \emph{global rules}. Curated excerpts of the four key prompts follow.

\subsection*{F.1\quad Global EvoForest Rules (system prompt, shared)}

\begin{promptbox}[Global Rules — excerpt]
GLOBAL EVOFOREST RULES & EVOLUTION PRINCIPLES

You reason about and evolve a "EvoForest": a population-encoded
program represented as a directed acyclic graph (DAG) of Python lambda
functions. Each node contains multiple alternative implementations.
Every valid configuration represents a distinct executable candidate.

1. EVOFOREST STRUCTURE (CONFIGURATION-BASED PARADIGM)
- A EvoForest is a DAG of nodes. Each node contains one or more
  alternatives, each implemented as a Python lambda function.

TWO KINDS OF NODES:
(A) INTERMEDIATE NODES: alternatives are COMPETING implementations.
    A "configuration" selects one alternative per intermediate node.
(B) OUTPUT NODE ("output"): alternatives are FEATURES. ALL output
    alternatives are evaluated for every configuration.
    Score = best config's performance across all configurations.

2. ENSEMBLE EVOLUTION OBJECTIVE
Build a diverse ensemble of complementary, high-quality predictors:
individually strong output features that capture different aspects
and combine well without redundancy. Each output alternative is an
expert in the ensemble; intermediate and callable nodes define
alternative ways those experts are built and combined.

3. QUALITY-DIVERSITY SEARCH DYNAMICS
Treat each alternative as a micro-program with ROLE, GENETIC LINEAGE,
PHENOTYPE IMPACT (statistics), and DESIGN PATTERN.
- **Exploit** strong alternatives (high max / mean).
- **Explore** weak or underrepresented alternatives.
- **Preserve diversity** by maintaining multiple distinct strategies.
THINK OUTSIDE THE BOX! Propose state-of-the-art algorithmic ideas,
not only parameter tweaks.

4. CODE-LEVEL CROSSOVER
1) Intra-node crossover: fuse strong alternatives of the same node.
2) Cross-node crossover: encapsulate recurring multi-step motifs.

5b. @globals -- PERSISTENT TRAINABLE PARAMETERS
The "@globals" node holds learnable tensor parameters that persist
across evolution steps. You may ADD new entries but NEVER modify or
remove existing entries (append-only). Supply an init expression;
the system wraps it in nn.Parameter and trains via backpropagation.
\end{promptbox}

\subsection*{F.2\quad Scientist Prompt (hypothesis generation)}

\begin{promptbox}[Scientist — system message]
You are an elite computational scientist and algorithm designer
specializing in algorithm evolution, representation learning, and
structural optimization. You analyze complex multi-path computational
graphs ("EvoForests"), diagnose structural bottlenecks, uncover
latent patterns, and propose high-value, state-of-the-art algorithmic
directions rather than incremental tweaks. THINK OUTSIDE THE BOX!

===================== TASK CONTEXT =====================
{task_context}   [one-time LLM-generated domain brief]
========================================================
\end{promptbox}

\begin{promptbox}[Scientist — user message (template)]
Generate 8-12 structured actionable hypotheses for the next step.
Do NOT output code or YAML. Be specific and grounded in diagnostics.

Each hypothesis must include:
- Hypothesis: a clear, actionable structural change.
- Rationale: grounded in the TOON diagnostics and graph structure.
- Expected Improvement: what metric or behavior should improve.
- Risk Mode: Conservative / Balanced / Risky.
- Self-Evaluation: Improvement (1-10), Creativity (1-10),
                   Implementability (1-10), Risk (1-10).

==== CURRENT EVOFOREST (YAML WITH STATS) ====
{annotated_yaml}

==== FEATURE DIAGNOSTICS (TOON -- PRIMARY EVIDENCE) ====
{diagnostics_toon}

==== EXPERIMENT LOG (OUTCOME HISTORY) ====
{island_memorandum}

[GLOBAL EVOFOREST RULES appended at end]
\end{promptbox}

\subsection*{F.3\quad Engineer Prompt (mutation synthesis)}

\begin{promptbox}[Engineer — user message (template)]
You are a world-class PyTorch engineer and computational systems
scientist. You critically evaluate the hypotheses in the system
context, extract the most meaningful ideas, discard weak or redundant
ones, and synthesize improved concepts when appropriate.

CONFIGURATION PARADIGM:
- Intermediate nodes have COMPETING alternatives. A "configuration"
  selects one alt per intermediate node; each config = an expert.
- Output node has FEATURES (all evaluated per config -> Ridge CV).
- Score = best config AUC. Adding output alts does NOT increase
  n_configs; adding intermediate alts multiplies them.

Your mutation must: maintain DAG validity (no cycles), improve
structural coherence or expressivity, align with the strongest
hypotheses, and remain computationally reasonable.

OUTPUT FORMAT -- YAML ONLY (MANDATORY)
remove:
  - some_alt_id
add:
  some_node:
    - "lambda args: <expr>"

==== CURRENT EVOFOREST (YAML WITH STATS) ====
{annotated_yaml}

==== EXECUTION ERRORS FROM PREVIOUS ATTEMPTS ====
{error_lines}
\end{promptbox}

\noindent\textbf{Composition.}\; The scientist receives the TOON, memorandum, and annotated graph; the engineer receives the annotated graph and execution errors in the user message, with the scientist's hypotheses and global rules in the system message. TOON and memorandum thus reach the engineer only through the scientist's hypotheses.

\subsection*{F.4\quad Memorandum Update Prompt}

\begin{promptbox}[Memorandum Update — system message]
You maintain an experiment log for one island of an evolutionary
EvoForest search. Record observations -- what happened, what's
noteworthy, what changed. No hypotheses, no recommendations.

You receive FEATURE DIAGNOSTICS (TOON) with each update. Don't copy
raw numbers -- instead note what STANDS OUT: anomalies, structural
issues, trends, and changes from the previous state.

TOON KEY -- how to read the diagnostics:
- ind_auc: standalone predictive power of this feature alone.
- imp: Ridge weight magnitude (sum-normalized).
- max_corr: highest |correlation| with any other feature.
- resid_corr / resid_sq: correlation with residual. High |resid_sq|
  = this feature captures signal the rest of the ensemble misses.
- effective_rank: how many independent signals the ensemble has.

Structure: [OUTCOME HISTORY] Running log of recent steps (last 5-8).
[STATE] Search progress, quality, diversity, landscape changes.
[WHAT WORKS] Patterns leading to ACCEPTED mutations.
[WHAT FAILED] Patterns leading to REJECTED or FAILED mutations.
[ERROR LOG] Implementation errors (accumulate, don't drop).

Rules: Under 500 words. No hypotheses. No markdown tables.
Never fabricate numbers -- only cite values from TOON or outcomes.
\end{promptbox}

\section{TOON Diagnostic Example}
\label{sec:toon}

The TOON report is the primary diagnostic artifact that the LLM receives. It contains a per-feature table for the best configuration's output features and a per-subnode table aggregating statistics across all configurations. Below is a truncated excerpt from the final global-best graph (8 of 55~features, 3 of 83~subnodes shown).

\begin{toonbox}
context:
  scoring: configuration-based (best config AUC = evoforest score)
  best_config_auc: 0.9577
  config_auc_range: [0.9351, 0.9577]
  global_ridge_auc: 0.9545
  fold_auc_std: 0.0012
  effective_rank: 38.7
  mean_max_corr: 0.825
  n_features_global: 3520
  n_features_best_config: 55
  n_configs: 64
  n_output_alts: 55
  n_clusters: 24
  n_nodes: 87
  max_depth: 7
  total_alternatives: 164
  multi_alt_nodes: 15
features[55]{name,depth,ind_auc,imp,sign,corr,max_corr,
    n_hi_corr,most_corr,cluster,cl_size,resid_corr,
    resid_sq,w_stab}:
  output_0,87,0.548,0.032,-,0.042,0.143,0,
    output_13,0,1,0.035,0.025,0.168
  output_3,87,0.582,0.011,+,0.157,0.508,0,
    output_36,2,1,0.025,-0.017,0.555
  output_5,87,0.555,0.013,-,0.030,0.589,0,
    output_18,4,1,0.062,0.015,0.151
  output_13,87,0.553,0.018,+,0.061,0.231,0,
    output_1,8,1,-0.033,-0.032,0.108
  output_17,87,0.902,0.027,-,0.131,1.000,4,
    output_24,10,5,0.273,-0.008,0.036
  output_20,87,0.888,0.086,-,0.105,0.998,1,
    output_37,12,2,0.233,0.057,0.015
  output_22,87,0.604,0.035,+,0.190,0.977,1,
    output_10,6,2,-0.007,0.010,0.061
  output_39,87,0.910,0.064,-,0.138,0.995,4,
    output_51,10,5,0.276,0.049,0.019
  ... (47 more features)
subnodes[83]{name,depth,n_deps,bttlnk,n_paths,
    cfg_auc_max,qi_max,qd_max,imp_max,
    ind_auc_max,resid_max}:
  @activation,0,0,T,3520,0.958,0.296,0.978,0.002,
    0.911,0.291
  cross_proj_attn,2,2,T,3520,0.958,0.296,0.978,0.002,
    0.911,0.291
  meta_gate,5,3,T,3520,0.958,0.296,0.978,0.002,
    0.911,0.291
  ... (80 more subnodes)
\end{toonbox}

\noindent The \texttt{features} table lets the LLM identify crown jewels, redundant-strong features, and weak-unique candidates. The \texttt{subnodes} table reveals which intermediate nodes participate in high-scoring configurations.

\section{Memorandum Example}
\label{sec:memorandum}

Each island maintains a persistent memorandum updated after every step. Below is Island~0's memorandum during late-stage search (AUC~$\approx$~0.958).

\begin{memobox}
[OUTCOME HISTORY]
- REJECTED: 0.957715 -> 0.957337 (D -0.000378) -- latest step.
- REJECTED: 0.957715 -> 0.948372 (D -0.009343).
- REJECTED: 0.957715 -> 0.957534 (D -0.000181).
- REJECTED: 0.957715 -> 0.956642 (D -0.001072).
- REJECTED: 0.957715 -> 0.957714 (D -0.000001).
- REJECTED: 0.957715 -> 0.956694 (D -0.001020).
- ACCEPTED: 0.957219 -> 0.957715 (D +0.000496).
- ACCEPTED: 0.955524 -> 0.957219 (D +0.001695).

[SNAPSHOT]
- Effective rank ~ 38.7 (<< 3520 features -> highly redundant).
- Mean max correlation ~ 0.825 (stable clustering).
- Best-config feature count = 55 (range 52-61).
- Graph size = 170 alternatives (net +8 recent).
- Acceptance rate (last 15) ~ 20% (down from ~27%).
- Island best AUC 0.9577; candidate 0.9573 (gap ~ +0.0035).
- Crown-jewel signals: output_0, output_13
  (high ind_auc, low max_corr, decent resid_sq).
- Redundant-strong cluster unchanged
  (outputs 17, 20, 34, 37, 39, 40, 51, 54).
- Weight-unstable: output_6 (w_stab 23.7),
  output_14 (68.7), output_41 (26.98).
- Residual-capturing: output_38, output_39, output_51.
- Weak + Unique: output_42 (low ind_auc, low max_corr).

[WHAT WORKS]
- Tiny norm tweaks to kan, var_gate, len_scale, attention,
  bilinear, shapelet banks remain safe (no crashes).
- AR3 mix (pre_ar3 + post_ar3 weighted by ar_gate)
  consistently adds ~ +0.0011 when isolated.
- kl_gauss addition contributed to an acceptance, stays stable.

[WHAT FAILED]
- Composite ridge (pre_var/(pre_var+post_var)) loses >0.003
  AUC and re-exposes ks_distance size mismatch.
- Large batches of tweaks across many modules break the graph.
- Manipulating weight-unstable features correlates with drops.
- Over-aggressive tweaks keep lowering scores.

[ERROR LOG]
- ks_distance size mismatch (tensor a (2499) vs b (999)).
- IndexError: list index out of range.
- Output[54] tensor size mismatch (pre_fft_power [10001,1250]
  vs post_fft_power [10001,500]).
- No new runtime errors in the latest step.
\end{memobox}

\noindent The memorandum compresses diagnostics and outcomes into a persistent narrative, enabling the LLM to track trends without re-analyzing raw data. Its hypothesis-free design prevents anchoring on stale recommendations.

\clearpage
\section{Transferable Discovery: Multi-Space Random Features}
\label{app:msrf}

A key promise of search-first ML is that computations discovered for one task can yield \emph{transferable algorithmic knowledge}.
This appendix describes a concrete example: a novel time-series classification method that originated from analyzing EvoForest's evolved graphs on the ADIA task.

\paragraph{Discovery origin.}
EvoForest consistently distributed its discovered computations across \emph{complementary representation spaces}---positional queries, global sequence-level projections, distributional summaries, and local convolutional patterns---rather than concentrating in any single feature family.
We distilled this motif into \textbf{Multi-Space Random Features (MSRF)}, a standalone framework that distributes random features across four spaces, each approximating a kernel over a different aspect of the signal:

\vspace{-2pt}
\begin{itemize}[leftmargin=*,topsep=2pt,itemsep=0pt]
\item \textbf{Temporal (TRF):} position-sensitive Gaussian-windowed projections, translation-\emph{variant} by design---complementary to translation-invariant convolutions.
\item \textbf{Global (GRF):} Random Fourier Features applied to the full sequence, compressed via a novel \emph{moment pooling} readout (mean and std of the embedding), yielding two features per projection matrix regardless of embedding dimension $D$.
\item \textbf{Statistical (SRF):} two-layer random projections of summary statistics (mean, variance, skewness, kurtosis, quantiles, autocorrelations), capturing distributional similarity independent of temporal ordering.
\item \textbf{Convolutional (CRF):} standard ROCKET-style \citep{dempster2020rocket} random convolutions for local pattern detection.
\end{itemize}
\vspace{-2pt}

\noindent The four types are approximately uncorrelated because they capture different invariances.
The combined representation $\Phi(\mathbf{x}) = [\phi_{\mathrm{temp}}, \phi_{\mathrm{glob}}, \phi_{\mathrm{stat}}, \phi_{\mathrm{conv}}]$ is classified by Ridge regression, which implicitly learns an optimal weighted kernel combination.

\paragraph{Results on the UCR archive (112 datasets).}
Under identical evaluation protocols (RidgeClassifierCV, standard train/test splits) against MiniRocket \citep{dempster2021minirocket} and the UCR archive \citep{dau2019ucr}:

\vspace{-2pt}
\begin{table}[H]
\centering
\small
\caption{MSRF on the 112-dataset UCR archive. W/T/L counts are relative to each baseline.}
\label{tab:msrf_ucr}
\vspace{-4pt}
\begin{tabular}{@{}lccl@{}}
\toprule
Comparison & Dims & Mean Acc. & W\,/\,T\,/\,L \\
\midrule
MSRF\textsubscript{full} & 1{,}410 & 0.808 & --- \\
MiniRocket & 10{,}000 & 0.794 & 59\,/\,14\,/\,39 \\
MiniRocket+MSRF & 10{,}410 & 0.803 & 67\,/\,31\,/\,14$^*$ \\
\bottomrule
\multicolumn{4}{@{}l@{}}{\footnotesize $^*$vs.\ MiniRocket alone}
\end{tabular}
\end{table}
\vspace{-6pt}

\noindent MSRF with 1{,}410 features achieves higher mean accuracy than MiniRocket with 10{,}000 features (0.808 vs.\ 0.794) using \textbf{86\% fewer dimensions}.
Appending just 410 non-convolutional features to MiniRocket improves accuracy on 67 datasets with only 14 losses (\textbf{83\% win rate}).
At a fixed budget of ${\sim}$1{,}400 features, multi-space allocation outperforms pure convolutions on 59\% of datasets.

\paragraph{Significance.}
The multi-space allocation principle---discovered autonomously through evolutionary search on a single competition dataset---generalized to a competitive standalone method across 112 diverse time-series domains, supporting the thesis that explicit computation discovery yields reusable algorithmic knowledge.

\smallskip
\noindent Code: \url{https://anonymous.4open.science/r/msrf/}

\end{document}